\documentclass[letterpaper, 10 pt, conference]{ieeeconf}  

\IEEEoverridecommandlockouts                              

\usepackage{times} 
\usepackage{amsmath} 
\usepackage[table, dvipsnames]{xcolor}
\usepackage{graphicx}
\usepackage{multicol}
\usepackage{multirow}
\usepackage{hyperref}
\usepackage{amssymb,amsfonts}

\hypersetup{
    colorlinks=true,
    linkcolor=red,
    urlcolor=blue}

\renewcommand{\vec}[1]{\boldsymbol{#1}}

\title{\LARGE \bf GelSlim 4.0: Focusing on Touch and Reproducibility}

\author{Andrea Sipos, William van den Bogert, Nima Fazeli%
\thanks{Supported by NSF CAREER \#2337870 and  NRI \#2220876.}%
\thanks{University of Michigan, Ann Arbor, MI, USA
        {\tt\small <asipos,willvdb,nfz>@umich.edu}}%
}

\begin{document}

\maketitle
\thispagestyle{empty}
\pagestyle{empty}

\begin{abstract}
    Tactile sensing provides robots with rich feedback during manipulation, enabling a host of perception and controls capabilities. Here, we present a new open-source, vision-based tactile sensor designed to promote reproducibility and accessibility across research and hobbyist communities. Building upon the GelSlim 3.0 sensor, our design features two key improvements: a simplified, modifiable finger structure and easily manufacturable lenses. To complement the hardware, we provide an open-source perception library that includes depth and shear field estimation algorithms to enable in-hand pose estimation, slip detection, and other manipulation tasks. Our sensor is accompanied by comprehensive manufacturing documentation, ensuring the design can be readily produced by users with varying levels of expertise. We validate the sensor’s reproducibility through extensive human usability testing. For documentation, code, and data, please visit the  \href{https://www.mmintlab.com/research/gelslim-4-0/}{project website}.
\end{abstract}

\section{INTRODUCTION}
Tactile sensing plays a crucial role in robotic manipulation, providing feedback that enhances a robot’s ability to interact with objects in dynamic environments. These sensors offer detailed information about contact interfaces, making them invaluable for tasks such as object handling, in-hand manipulation, and force regulation. Given their potential to significantly improve robotic perception and control, tactile sensors have been a focus of research for decades, with many designs pushing the boundaries of what is possible in terms of sensing capabilities. However, despite this progress, the widespread integration of tactile sensors into full-stack robotic systems remains limited.

\begin{figure}
    \centering
    \includegraphics[width=\linewidth]{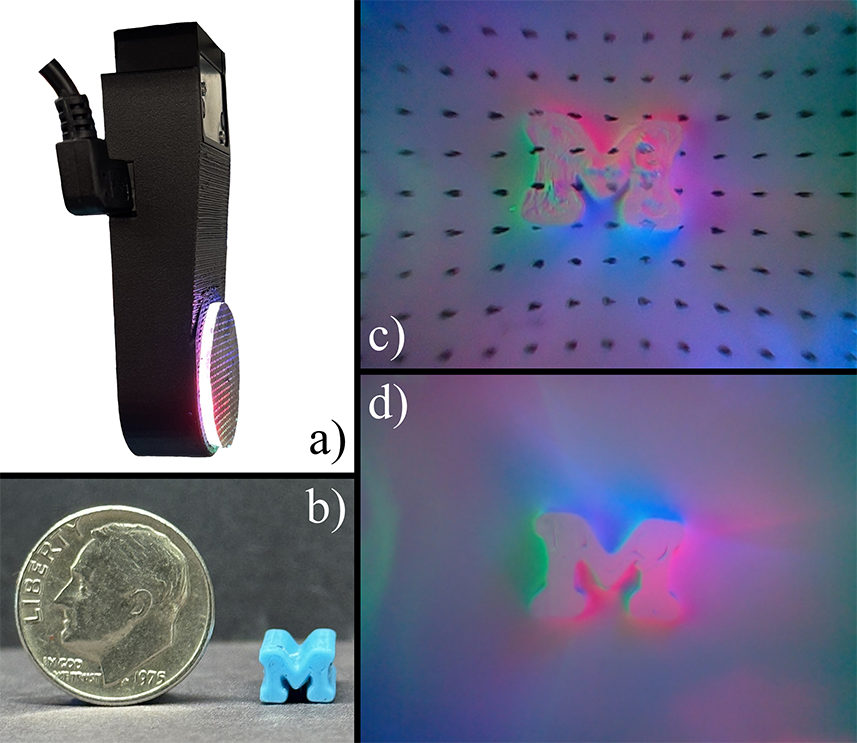}
    \caption{a) GelSlim 4.0 with shear tracking markers mounted on a WSG-50 gripper. b) 3D printed Block M next to a dime for scale. c) Block M indented into a GelSlim 4.0 with shear tracking markers. d) Block M indented into a GelSlim 4.0 without shear tracking markers.} \vspace*{-6mm}
    \label{fig:teaser}
\end{figure}

One of the key barriers to this integration is that much of the research has prioritized advancing the capabilities of these sensors rather than addressing their reproducibility and manufacturability. While the push for higher resolution, sensitivity, and multi-modal data capture has driven innovation, it has also resulted in designs that are often complex and difficult to build, especially for individuals who are not specialized in hardware development. This complexity can be a significant hurdle, particularly in environments where sensors are subject to wear and tear and require frequent replacement or maintenance. For tactile sensing to scale in real-world applications, the community must consider not only the sophistication of sensor designs but also their practicality, including how easily they can be produced, modified, and integrated into diverse robotic systems by researchers, educators, and students alike.

In this paper, we address these challenges by presenting GelSlim 4.0, a new vision-based tactile sensor focused on reproducibility, affordability, and accessibility. Our contributions include a customizable finger and easily manufacturable lens, both designed to significantly reduce the cost and complexity of building the sensor. We provide an open-source code base and dataset for depth and shear field estimation, which are fundamental capabilities of visuotactile sensors. This dataset includes GelSlim 4.0 images from multiple sensors and multiple objects. The sensor is accompanied by an extensive user-tested manufacturing manual complete with video documentation, ensuring that users at all levels can easily produce and modify the sensor. This effort explicitly lowers the barrier to entry for robotics researchers who may primarily focus on algorithms rather than hardware, as well as educators seeking to introduce tactile sensing into STEM curricula. By making the sensor both affordable and easy to build, we aim to democratize the use of tactile sensing in robotics research and education, empowering a broader range of users to incorporate this powerful modality into their work.
\section{RELATED WORK}
\label{sec:rel_work}
There are many different types of proprioceptive and tactile sensors that are commercially available or open-source that provide sensory information from contact-rich interactions. Some of these proprioceptive sensors include force sensors, force-torque sensors, and force-torque sensor arrays. Within tactile sensing, there are several different styles of sensors such as robotic skin \cite{fan2022enabling, bimbo2016hand}, bio-mimetic sensors \cite{wettels2014biotac}, and visuotactile sensors \cite{kuppuswamy2020softbubble, yuan2017gelsight, lepora2022digitac, yamaguchi2019tactile} that are used in robotic manipulation. 

In comparison to proprioceptive sensors, there are very few high-quality tactile sensors available commercially. The tactile sensing solutions available can be broadly categorized as robotic skins/taxel-based sensors and visuotactile sensors. 

\subsection{Robotic Skins and Taxel-Based Sensors} Robotic skins attempt to emulate the sensory capacity of human skin, which uses several different types of mechanoreceptors to sense vibration, pressure, stretch, etc. In a human hand, it has been estimated that there are approximately 17,000 tactile units \cite{vallbo1999mechanoreceptors}. Achieving this level of sensor density and parsing the information are still open problems in robotic skin design. While there are no commercially available full-robot skin systems at this point in time to the authors' knowledge, there are several commercially available sensors that focus on the fingertip or palm of robotic grippers. These systems use several ``taxels'', or single tactile sensory units, to span sensory areas. Some of these available sensors include \href{https://contactile.com/products/}{PapillArray and Force Button} from Contactile, \href{https://www.xelarobotics.com/tactile-sensors}{uSkin sensors} from XELA Robotics, and force-sensitive resistor technology used by companies like \href{https://www.psyonic.io/}{Psyonic} in their Ability Hand. 

\subsection{Visuotactile Sensors} Visuotactile sensors are different from robotic skins or taxel-based sensors in that they use cameras to observe the contact surface at a much higher resolution. Visuotactile sensing is a powerful tool in robotic manipulation that enables foundational skills for dexterous manipulation such as estimating and controlling extrinsic contacts \cite{ma2021extrinsic, kim2022extrinsic, oller2022membranes} and in-hand object poses \cite{villalonga2021tactile, bauza2023tac2pose, kim2024texterity}. These visuotactile methods often perform best when the object renders a richly-featured tactile imprint. 

There are four main visuotactile sensors for robotics available for purchase: 
\href{https://www.gelsight.com/product/gelsight-mini-system/}{GelSight Mini}, \href{https://www.gelsight.com/product/gelsight-mini-robotics-package/}{GelSight Mini Robotics Package}, \href{https://www.gelsight.com/product/digit-tactile-sensor/}{DIGIT}, and \href{https://www.fingervision.jp/en/service-product}{FingerVision}. 

The GelSight Mini Robotics Package differs from the GelSight Mini and DIGIT in that it includes tracking markers, which aid in shear and slip estimation. The GelSight Mini and DIGIT alone do not have these markers. The GelSight Mini costs \$499 with \$49 replacement gels, the GelSight Mini Robotics Package costs \$549 with \$69 replacement gels, and the DIGIT Tactile Sensor costs \$350 with \$40 replacement gels at the time of writing. 

While not many tactile sensors are commercially available, there are several that have open-source resources so that interested users can produce them, including the DIGIT \cite{lambeta2020digit}, GelSlim 3.0 \cite{taylor2022gelslim}, 9DTact \cite{lin20239dtact}, and Soft Bubble Gripper \cite{kuppuswamy2020softbubble}. 

These resources are tremendously valuable to the robotics community because producing sensors in-house often drives down cost and allows more researchers to access them. For example, the materials to produce a single GelSlim 3.0 sensor total approximately \$122. This is almost five times less than the cost of the commercially available GelSight Mini Robotics Package. It is important that the robotics community has access to high-quality open-source tactile sensing options so that more research labs and educators can access these powerful tools in their research and coursework.

However, existing open-source visuotactile sensors have various limitations. The GelSlim 3.0 is unmaintained and has critical components that have been discontinued by the manufacturer. The DIGIT and 9DTact are intended to be used with anthropomorphic hands and consequently have smaller sensing areas and lack adaptors for common parallel-jaw grippers. The Soft Bubble Gripper utilizes a highly compliant membrane which results in large relative motions between grasped objects and the gripper itself. 

There are many more sensor designs that have been published but are not fully open-source or commercially available, including but not limited to: EyeSight Hand \cite{romero2024eyesight}, GelLink \cite{ma2024gellink}, ROMEO Fingers \cite{liu2024romeo}, RainbowSight \cite{tippur2024rainbowsight}, TacTip \cite{chorley2009tactip}, ViTacTip \cite{zhang2024vitactip}, GelSight 360 \cite{tippur2023gelsight360}, GelSight Svelte/GelSight Svelte Hand \cite{zhao2023svelte}, GelSight Fin Ray \cite{lui2022finray}, GelSight Baby Fin Ray \cite{liu2023bbfinray}, StereoTac \cite{roberge2023stereotac}, See-Through-Your-Skin Sensor \cite{hogan2021sts}, Fingervision, GelStereo \cite{cui2022gelstereo}, GelStereo Palm \cite{hu2023palm}, GelStereo BioTip \cite{cui2023biotip}, HaptiTemp \cite{abad2022haptitemp}, and TIRgel \cite{zhang2023tirgel}. Several of these sensors have materials and methods that are similar to our GelSlim 4.0 sensor but do not have the in-depth open-source documentation for production that we provide. Without these resources, novel sensor designs can be difficult for other researchers to recreate and use in their own work. The key contribution of this work is that we open-source our resources and user-test them with novice users from various backgrounds to ensure reproducibility. We believe that this will enable researchers to coalesce around a visuotactile sensing platform, which will in turn allow for direct comparisons and benchmarking between novel algorithms for dexterous manipulation. Similar work has been very impactful in other areas of robotics, such as the Open Source Leg in powered prosthesis research \cite{azocar2020osl}.
\section{METHODOLOGY}
We separate our contributions into three areas. First, we present the GelSlim 4.0 to address several limitations of the GelSlim 3.0. Second, we provide open-source algorithms for depth estimation and shear field estimation for the GelSlim 4.0 sensor. Third, we conduct a human study with 17 novice users from various backgrounds to examine the reproducibility of critical GelSlim 4.0 components. 

\begin{figure}
    \centering
    \includegraphics[width=\linewidth]{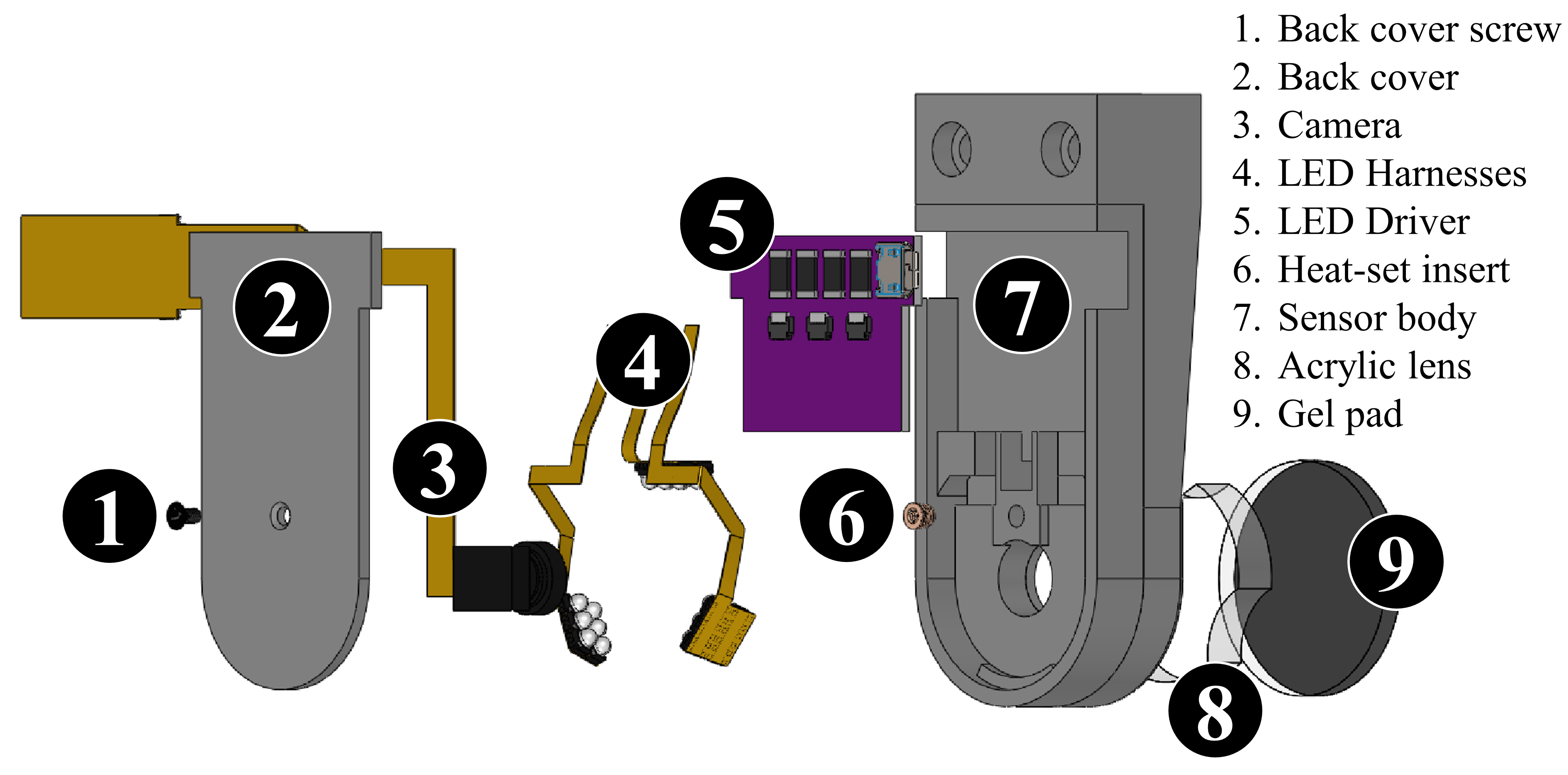}
    \caption{Exploded diagram of the GelSlim 4.0 for the WSG-50 Gripper.}
    \label{fig:exploded_model} \vspace*{-6mm}
\end{figure}

\begin{figure*}
    \centering
    \includegraphics[width=\linewidth]{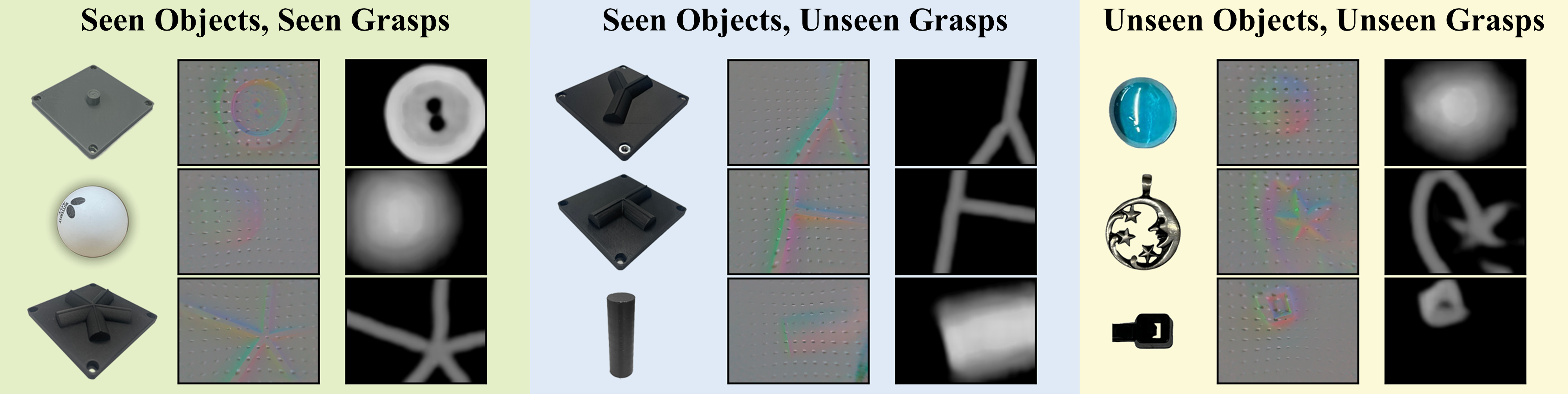}
    \caption{Results of depth estimation using our proposed method. The rows in each section consist of a photo of the real object, the distorted RGB difference image obtained from the GelSlim 4.0 sensor, and the estimated rectified depth image. \textit{Left:} 3D printed button, ping pong ball, 3D printed asterisk shape. \textit{Middle:} 3D printed Y shape, 3D printed T shape, 3D printed cylindrical peg. \textit{Right:} Marble, necklace pendant, small zip tie.} \vspace*{-6mm}
    \label{fig:depth_recon}
\end{figure*}

\subsection{GelSlim 4.0 Updates}
\label{sec:gelslim_updates}
The GelSlim 3.0 is an important visuotactile sensor for which key algorithms have been developed, as discussed in Sec.~\ref{sec:rel_work}. However, the sensor has several limitations. Among them is that the sensor is no longer maintained. Another significant limitation is that the CAD files provided in the GelSlim 3.0 resources have limited capacity for modification. This makes it difficult to modify the sensor for new end-effectors, tweak tolerances to meet fabrication capabilities, or control properties like stiffness. A critical component of the GelSlim 3.0 has also been discontinued by the manufacturer without replacement, necessitating these difficult modifications. Finally, the shaping lens used in the GelSlim 3.0 is treated as consumable, but it is the most expensive of the sensor components. These issues have created challenges for the community in building and using this sensor as intended. In order to revitalize the sensor, we introduce several major modifications in the GelSlim 4.0 sensor:
\begin{itemize}
    \item Comprehensive written and video documentation of GelSlim 4.0 manufacturing processes
    \item Increased customizability via detailed open-source design files available in both Solidworks and OnShape
    \item Consolidated sensor design for ease of assembly
    \item Redesigned consumable lens that dramatically reduces manufacturing time and cost
\end{itemize}
These changes also increase access to the sensor within the robotics community. Access to the underlying design trees in the CAD files enables rapid prototyping across various grades of equipment from hobbyist to professional. Releasing the files in OnShape gives anyone with a .edu email address access to those features as well, rather than restricting to those with Solidworks licenses.  

A tangle of issues with the GelSlim 3.0 stems from a small, discontinued blue LED whose optical properties are used to optimize the shaping lens. The depth estimation capabilities of the GelSlim 3.0 were built around photometric stereo, which makes the assumption of illumination homogeneity on the contact surface. Consequently, depth estimation is degraded in any non-homogeneous areas. To achieve homogeneous illumination, the GelSlim 3.0 used a complex shaping lens to diffuse light across the contact surface. Geometric parameters of the lens were optimized in a cumbersome two-step process between Solidworks and LightTools. This optimization considered material properties of the acrylic and optical properties of the LEDs. The resulting high-precision lens costs \$50+ (when purchased 6 at a time from a turnkey fabrication service) because it requires CNC machining and a transparent finish. Because this shaping lens was optimized for a specific set of LEDs, it is not robust to changes in LED properties. The closest blue LED available to the discontinued LED has a significantly narrower viewing angle, which resulted in dark spots on the sensor contact surface and an inability to use photometric stereo for precise depth estimation.

These cascading impacts led us to simplify the lens design and switch to a learned depth estimation method (discussed in Sec.~\ref{sec:gelslim_algs}) to facilitate future iterations when part availabilities inevitably change. The new lens design (shown in Fig.~\ref{fig:exploded_model}) is planar, meaning that it can be laser cut on standard equipment. This reduction in complexity means the lens can be fabricated in minutes for a mere \$0.13. The nearly 400x reduction in cost for this component also allows it to be treated as truly consumable. This is extremely important because a fundamental limitation of tactile sensors is that their contact interfaces fatigue with time and use. With the GelSlim 4.0, it becomes cheap, quick, and easy to replace these components as they wear out. These changes bring the cost of the GelSlim 4.0 down from $\sim$\$122 to $\sim$\$65.

\subsection{GelSlim Algorithms}
\label{sec:gelslim_algs}
To accompany the sensor refresh, we provide open-source Python3 implementations for the GelSlim 4.0 that enable key tactile sensing capabilities: depth estimation and shear field estimation. We provide these implementations to demonstrate that despite the significant changes to the sensor discussed in the previous section, the GelSlim 4.0 maintains its utility for dexterous manipulation. 

\subsubsection{Depth Estimation} The GelSlim 3.0 was accompanied by an analytical depth estimation method based on photometric stereo and sensor calibration \cite{taylor2022gelslim}. This method required calibrating each new sensor. As we discussed in Sec.~\ref{sec:gelslim_updates}, the GelSlim 3.0 used an expensive, optimized shaping lens to achieve the required illumination homogeneity of the contact surface. By refactoring the shaping lens, the GelSlim 4.0 removes a high-cost component at the expense of this analytical depth reconstruction method. Here, we present a learning-based approach for depth reconstruction in its place. In our approach, we train a U-Net \cite{ronneberger2015unet} to predict an RGB-to-depth mapping.

\subsubsection{Shear Field Estimation} The optional dot pattern on the gel pads of the GelSlim 4.0 sensor enables optical flow-based tracking of dot displacements using standard OpenCV functions. These displacements are used to generate vector fields in pixel space. This shear field representation of tactile information has proven useful for force-rich manipulation tasks and slip detection \cite{zhang2018fingervision, dong2020cables}. Our open-source repository for this algorithm can operate on either distorted or rectified sensor images. Given a function $f$ that calculates optical flow between two RGB images we can generate a discrete shear field $\vec{u}$ sampled on an $H \times W$ grid. With this function we can find two approximations for the shear field:
\begin{align*}
&\text{1. Using time-adjacent images: } & \vec{u}_t &\approx f(\mathbf{I}_t,\mathbf{I}_{t-1})+\vec{u}_{t-1} \\ &\text{2. Using undeformed image $\mathbf{I}_0$: } & \vec{u}_t &\approx f(\mathbf{I}_t,\mathbf{I}_0) \label{eq:shear_approx}
\end{align*}
Approximation 1 can suffer from integration drift, and Approximation 2 can suffer from inaccurate registration between the deformed and undeformed markers if displacement magnitudes are high. Our implementation offers both of these approximations alongside a weighted approximation based on the displacement measurement $d$:
\begin{align*}
\vec{u}_t &= \omega_t(f(\mathbf{I}_t,\mathbf{I}_{t-1})+\vec{u}_{t-1})+(1-\omega_t)f(\mathbf{I}_t,\mathbf{I}_0)\\ \omega_t &= \frac{1}{1-\text{exp}(-k(d_t-m))}\\
d_t &= \text{max}(\text{std}(\vec{u}_{t_x}),\text{std}(\vec{u}_{t_y}))
\end{align*}

\begin{figure}
    \centering
    \includegraphics[width=0.8\linewidth]{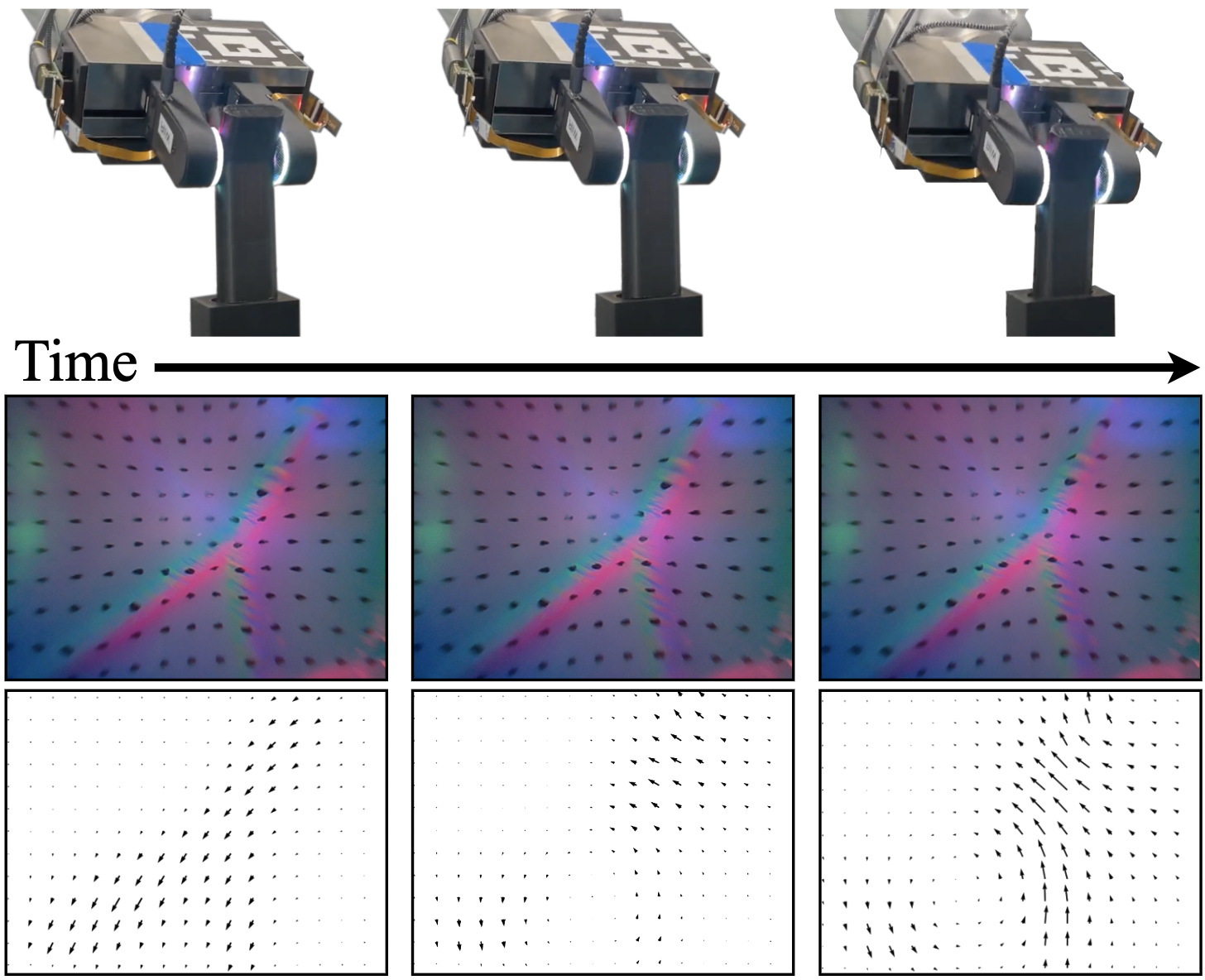}
    \caption{Distorted RGB images and their shear fields obtained while a robot performs a peg insertion task: first contact, hole exploration, and insertion.} \vspace*{-6mm}
    \label{fig:shear}
\end{figure}

\subsection{Human Study on GelSlim 4.0 Reproducibility}
\label{sec:human_study}
Providing comprehensive instructions for fabricating a visuotactile sensor to novice users is challenging for several reasons. One major challenge is that the anticipated users are roboticists, a notably interdisciplinary group. Consequently, it is difficult to predict what the manufacturing ability of a given user will be. It is therefore crucial to conduct a usability study with a broad audience to identify pain points and correct them before releasing open-source instructional resources to the community. 

\subsubsection{Experimental Design}
We select three tasks for this study from the full GelSlim 4.0 manufacturing instructions:
\begin{itemize}
    \item Task 1: Soldering an LED Harness
    \item Task 2: Polishing a lens
    \item Task 3: Assembling GelSlim 4.0 from subcomponents
\end{itemize}
A labeled diagram showing each of the core components is shown in Fig.~\ref{fig:exploded_model}. We select these tasks because they are critical to the successful fabrication of the sensor. We choose the soldering task because surface mount device (SMD) soldering can be an unfamiliar and intimidating skill that may dissuade new users from making the sensor. Next, we choose the polishing task because the quality of the lens heavily impacts the final sensor images. Finally, we select the assembly task because, while it is possible to pay for turnkey fabrication of GelSlim 4.0 components, it is impossible to purchase a fully assembled sensor.

Task 1 and Task 3 have binary metrics for task success. In Task 1, all 6 LEDs on the LED Harness must light in the correct orientation for Task 1 to be successful. In Task 3, there are three metrics for task success. First, the sensor is connected to power and all 3 LED Harnesses must light. Second, the sensor must have the gel and lens assembly properly installed. Third, the ribbon cable for the camera must exit the sensor body via the correct outlet. All three metrics must be successful for Task 3 to be successful. Task 2 is qualitatively evaluated for success by the facilitator. If the lens is sufficiently smooth, Task 2 is successful.

We ran 4 versions of the instructions in this study with 17 total participants: V0 ($n=1$), V1 ($n=3$), V2 ($n=3$), and V3 ($n=10$). In between each version we made updates to the instructions based on our observations of the participants as well as their feedback.

We use NASA Task Load Index (TLX) \cite{hart1986nasa} surveys to evaluate subjective workload for each of the tasks. The NASA-TLX measures workload along six dimensions: mental demand, physical demand, temporal demand, performance, effort, and frustration. NASA-TLX provides a score from 0 to 100, where a lower score indicates lower workload. 

\subsubsection{Participants} 
All participants were 18 years or older, able to fully use both arms and hands, able to communicate verbally in English, able to read and write in English, and self-identified STEM researchers, educators, and/or mentors. Demographic information about age, gender, race, etc. was not collected. Participants gave informed consent in accordance with University human participant protection policies. Participants were not compensated for their participation. The study was approved by the University of Michigan’s Institutional Review Board (HUM00253433).

\subsubsection{Procedure}
Upon arrival, participants were given an overview of the study and asked to fill out a pre-survey. Then, they received instructions for the NASA-TLX survey. In the task portion of the study, the procedure for each task was the same. First, the task was introduced at a high level. Next, participants donned proper personal protective equipment (PPE) for the task and were reminded that the facilitator would only be able to answer questions about safety, not about the provided instructions. Then, instructional resources were provided, video recording of the workspace was started, and participants were prompted to begin the task. 

For Tasks 1 and 2, participants were provided with a labeled diagram of the workspace, written instructions, and a video. For Task 3, only a labeled diagram of the workspace and written instructions were provided. It is important to note that instructions for checking work were not included for Task 1 and Task 3. While these instructions do exist in the released version, they were not included in this study to minimize the time and materials required for each participant by limiting them to one attempt at each task. After each task, participants were instructed to doff their PPE and immediately begin the NASA-TLX survey for that task. After all 3 tasks were completed, instructional materials were returned to participants and they completed a post-survey where they had the ability to offer specific long-form feedback on the resources they used to complete the tasks. 
\begin{table}
    \centering
    \begin{tabular}{c c c c c c}
        \hline
        \multirow{2}{*}{\textbf{Task}} & \multicolumn{4}{c}{\textbf{Success Rate}} & \multirow{2}{*}{\textbf{Overall}} \\
        \cline{2-5}
        & V0 & V1 & V2 & V3 & \\
        \hline
        1 & 0/1 & 2/3 & 3/3 & 8/10 & 13/17 \\
        2 & 0/1 & 3/3 & 3/3 & 9/10 & 15/17 \\
        3 & 1/1 & 3/3 & 3/3 & 9/10 & 16/17 \\
        \hline
    \end{tabular}
    \caption{Task success across instruction versions}\vspace*{-10mm}
    \label{tab:success}
\end{table}

\begin{table*}
    \centering
    \begin{tabular}{ r  c  c  c  c }
        \hline
        \multirow{2}{*}{\textbf{Statement}} & \multicolumn{2}{c}{\textbf{Difference (Post - Pre)}} & $\textbf{H}_{\mathbf{a^{+}}}$: $\mu_{\text{pre}} < \mu_{\text{post}}$ & $\textbf{H}_{\mathbf{a^{-}}}$: $\mu_{\text{pre}} > \mu_{\text{post}}$ \\
        \cline{2-5}
         & Mean & Standard Deviation & $p$-value & $p$-value \\
        \hline
        I can work with tools and use them to build things & 0.1 & 0.568 & 0.296 & - \\
        \rowcolor{ForestGreen!25} \textbf{I can manipulate small components} & \textbf{0.3} & \textbf{0.483} & \textbf{0.041} & - \\
        I can assemble things & 0.0 & 0.0 & - & - \\
        \rowcolor{ForestGreen!25} \textbf{I can fabricate electronic components} & \textbf{0.8} & \textbf{0.919} & \textbf{0.011} & - \\
        I like building things & -0.1 & 0.316 & - & 0.172 \\
        I prefer to make things from scratch & -0.2 & 0.422 & - & 0.084 \\
        I prefer to buy pre-made things & -0.1 & 0.316 & - & 0.172 \\
        I like learning new skills & -0.1 & 0.316 & - & 0.172 \\
        I am easily frustrated & 0.0 & 0.471 & 0.500 & 0.500 \\
        I am patient & 0.2 & 0.421 & 0.084 & - \\
        \hline
    \end{tabular}
    \caption{Paired one-tailed t-test results for each statement from the pre- and post-surveys responses of V3 participants. Statements with significant change at the $p < 0.05$ significance level are bold and highlighted.}\vspace*{-10mm}
    \label{tab:ttests}
\end{table*}

\section{EXPERIMENTS AND RESULTS}
\subsection{GelSlim Algorithms}
\subsubsection{Depth Estimation}
To train the U-Net for RGB-to-depth mapping, we generated ground-truth depth images for a variety of grasped objects from CAD models. In total, we used 15 objects with known geometry in the training and validation data. We collected RGB images using GelSlim 4.0 sensors as the fingers of a Weiss Robotics WSG-50 gripper mounted on a Kuka LBR iiwa R820 robot arm. We used data from 4 sensors with shear tracking markers. The objects were fixed to a known location in the environment. We used the Kuka's proprioception and each object's CAD model to generate a point cloud of the grasped object's surface. We assume that this point cloud is well-aligned with the GelSlim 4.0 camera frame. For each object, this resulted in 200 data points used for training. Each data point consists of a distorted RGB difference image (the subtraction of the deformed and undeformed images) and its paired rectified depth image. For the objects used in training, 10\% of the data was reserved for validation. 5 new objects were also reserved for validation during training. Additionally, 3 test objects were completely unseen. In Fig.~\ref{fig:depth_recon} we show the measured distorted RGB difference images and the resulting estimated rectified depth images for samples of seen objects/seen grasps, seen objects/unseen grasps, and unseen object/unseen grasps.

\subsubsection{Shear Field Estimation}
Our shear field algorithm extracts an $H=13 \times W=18$ grid of 2D shear vectors that form a 2-channel image. This image can be used with CV algorithms. The values of $H$, $W$, and other parameters are configurable. For more details, visit the \href{https://www.mmintlab.com/research/gelslim-4-0/}{project website}. Examples of the shear field using the weighted approximation described in Sec.~\ref{sec:gelslim_algs} are shown in Fig.~\ref{fig:shear}.

\subsection{Human Study on GelSlim 4.0 Reproducibility}
\subsubsection{Participant Demographics}
\begin{figure}
    \centering
    \includegraphics[width=\linewidth]{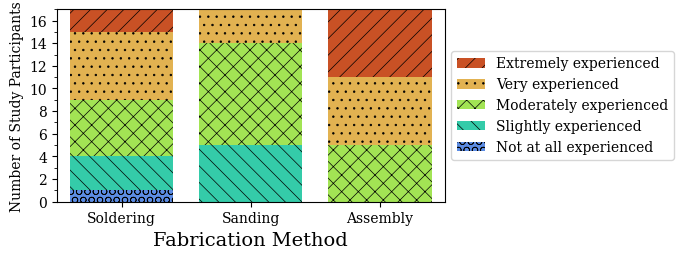}
    \caption{Prior experience of study participants across all versions} \vspace*{-6mm}
    \label{fig:prior_experience}
\end{figure}

In the demographics portion of the pre-survey, participants self-reported their undergraduate fields of study as including chemical engineering, chemistry, computer science, mathematics, computer engineering, mechanical engineering, robotics, and electrical engineering. Where applicable, participants self-reported their graduate fields of study as including chemical engineering, polymer science and engineering, computer science, robotics, and mechanical engineering. Participants self-evaluated their prior experience with manufacturing methods related to the GelSlim 4.0 manufacturing instructions on a 5-point Likert scale. The results of this survey are shown in Fig.~\ref{fig:prior_experience}. Approximately half of the participants in this study verbally expressed to the facilitator during or after the study that it was their first time SMD soldering. 

\subsubsection{Task Success and Workload Evaluation}
\begin{figure}
    \centering
    \includegraphics[width=\linewidth]{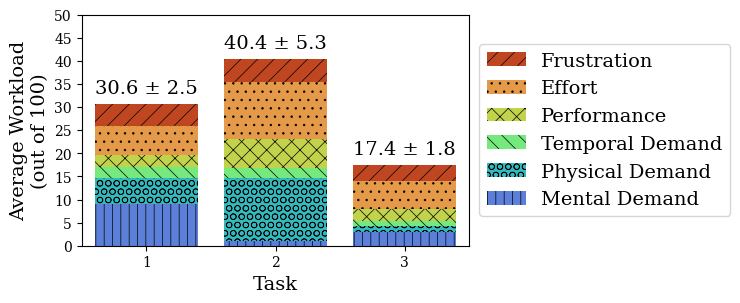}
    \caption{Average workload composition of each task for V3 participants.} \vspace*{-6mm}
    \label{fig:workload_composition}
\end{figure}

Task success across instruction versions is shown in Table~\ref{tab:success}. Across all tasks, both the failure cases and the success cases with high workloads provided insight to how the manufacturing resources could be improved. The magnitude and composition of these workload scores is shown in Fig.~\ref{fig:workload_composition}.

There were 4 failure cases in Task 1:
\begin{enumerate}
    \item V0: The participant flipped all 6 LEDs.
    \item V1: The participant did not know how to properly clean the solder stencil between attempts to spread solder paste. The final circuit was shorted due to large amounts of excess solder on the board.
    \item V3: The participant turned off the hot plate before the solder paste fully flowed when they saw the board's long tail curling, even though curling is allowable.
    \item V3: The participant flipped 1 of 6 LEDs. 
\end{enumerate}

In each of these cases, changes were made to the instructions to be more explicit and address the points of confusion that caused failure. In addition to the participants in the first and fourth failure cases, there were also several successful participants who did not use the provided multimeter to identify LED orientation. Instead, they tried to visually identify the orientations of LEDs without a magnifying device. This task showed a higher mental demand than either of the other tasks, and we hypothesize that this is because some participants didn't have access to the tools they wanted to use to determine LED orientation. An alternative method using a magnifying glass was added to the instructions after the study concluded.  

In Task 2, there were 2 failure cases. These participants expressed that they had difficulty understanding the end conditions for the task. This sentiment was echoed by other participants who, while successful, reported high effort scores and expressed that they couldn't tell when to stop sanding based on the provided resources. One challenge in communicating the end condition is that the lens is clear; consequently, it is difficult to capture its surface quality. To address this, surface defects have been annotated on high-quality photos of several lenses in the released instructions. Another difficulty was that participants did not understand how the quality of the lens impacts the final image quality of the sensor. Sensor images from lenses with various defects have been added to the released instructions to address this.

The lone failure case from Task 3 did not result in updates to the manufacturing resources. In this case, a participant failed to fully insert one LED Harness into its connector. No changes were made to the instructions to account for this failure case because the study purposely did not provide participants the ability to check their work. Two participants who successfully completed Task 3 expressed high levels of frustration, mental demand, and effort that resulted in high overall workload scores. These participants expressed wanting to use tweezers, which were not supplied in this task, to connect electrical components. Other than these two cases, Task 3 had a relatively uniform distribution of task workload that amounted to a low overall workload score.

Finally, we examined prior relevant experience as a predictor for workload scores. We performed linear regressions of task workload vs. relevant prior experience for each task in V3 and found low correlation between them. The $R^2$ values for Tasks 1, 2, and 3 were 7.04\%, 24.21\%, and 1.29\% respectively. This result shows that prior experience or lack thereof is not strongly correlated with overall workload when using our manufacturing resources to produce critical sensor components. 

\subsubsection{Participant Confidence and Attitudes}
In the pre- and post-survey, participants were asked to rate their agreement with the statements listed in Table~\ref{tab:ttests} using a 5-point Likert scale (1=Strongly Disagree to 5=Strongly Agree). We compare V3 participant responses to these statements in the pre- and post-surveys using paired one-tailed t-tests to evaluate whether there were any changes in confidence or attitudes as a result of completing the tasks in this study. To determine the direction of the tests, we use the mean difference of the scores for each statement (post$-$pre). We use the following null and alternative hypotheses to test each statement:

\begin{center}
$\textbf{H}_{\mathbf{0}}$: $\mu_{\text{pre}} = \mu_{\text{post}}$
\\
$\textbf{H}_{\mathbf{a^{+}}}$: $\mu_{\text{pre}} < \mu_{\text{post}}$, 
$\textbf{H}_{\mathbf{a^{-}}}$: $\mu_{\text{pre}} > \mu_{\text{post}}$
\end{center}

If the mean difference was positive, we used $\textbf{H}_{\mathbf{a^{+}}}$. If the mean difference was negative, we used $\textbf{H}_{\mathbf{a^{-}}}$. We found statistically significant change ($p < 0.05$) in responses to two statements: ``I can manipulate small components" and ``I can fabricate electronic components". Both of these changes were increases, indicating that participant confidence increased after completing the tasks in this study. 
 
\section{CONCLUSION}
In this paper, we contribute a much-needed update to the popular GelSlim 3.0 visuotactile sensor. In the GelSlim 4.0 design, we address several key limitations of the GelSlim 3.0, including its lack of customizability and the cascading effects of part availability on sensor design. We additionally provide implementations of depth and shear field estimation on the GelSlim 4.0, enabling these fundamental capabilities despite the extensive changes to the sensor design. Finally, we provide comprehensive written and video documentation for fabricating, mounting, and using the GelSlim 4.0 and verify the reproducibility of key sensor components via a human study consisting of novice users from a variety of backgrounds. The goal of this work is to provide roboticists with an accessible visuotactile sensing platform, and we hope to see adoption and community contributions to this platform in the near future. 
\section{ACKNOWLEDGMENT}
The authors would like to thank the following researchers for their contributions to this work: Samanta Rodriguez, Hannah Baez, Abigail Rafter, and Japmanjeet Singh Gill. 
\clearpage
\bibliographystyle{IEEEtran}
\bibliography{IEEEabrv,ref}
\end{document}